\documentclass{article}
\usepackage[margin=1.5in]{geometry}

\usepackage{url}  
\usepackage{graphicx}  
\usepackage{listings}
\usepackage{booktabs}
\usepackage{amsmath}
\usepackage{multicol}
\usepackage{todonotes}
\usepackage{enumitem}
\usepackage{float}

\date{}

\title{Learning Multi-Party Turn-Taking Models \\from Dialogue Logs}

\usepackage{authblk}
\author{Maira Gatti de Bayser}
\author{Paulo Cavalin}
\author{Claudio Pinhanez}
\author{Bianca Zadrozny}
\affil{IBM Research, Rio de Janeiro, Brazil}

\begin{document}

\label{firstpage}
\maketitle

\begin{abstract}
This paper investigates the application of machine learning (ML) techniques to enable intelligent systems to learn multi-party turn-taking models from dialogue logs. The specific ML task consists of determining who speaks next, after each utterance of a dialogue, given who has spoken and what was said in the previous utterances. With this goal, this paper presents comparisons of the accuracy of different ML techniques such as Maximum Likelihood Estimation (MLE), Support Vector Machines (SVM), and Convolutional Neural Networks (CNN) architectures, with and without utterance data. We present three corpora: the first with dialogues from an American TV situated comedy (chit-chat), the second with logs from a financial advice multi-bot system and the third with a corpus created from the Multi-Domain Wizard-of-Oz dataset (both are topic-oriented). The results show: (i) the size of the corpus has a very positive impact on the accuracy for the content-based deep learning approaches and those models perform best in the larger datasets; and (ii) if the dialogue dataset is small and topic-oriented (but with few topics), it is sufficient to use an agent-only MLE or SVM models, although slightly higher accuracies can be achieved with the use of the content of the utterances with a CNN model.
\end{abstract}

\section{Introduction}

Conversational systems have entered mainstream applications in recent years, as a result, we have seen a huge number of dyadic chatbots becoming available to users, from chit-chat bots which can converse about generic topics to entertain people, to chatbots which are experts and can provide useful services through natural language \cite{Serban17} \cite{Budzianowski18}, such as booking trains, flights or controlling lights. 
However, in everyday life, and particularly in the case of \textit{smart speakers}, such as \textit{Echo} and \textit{Google Home}, it is common that the conversational systems are in the presence of multiple users which not only talk to the chatbot but also to each other. For a chatbot to determine whether and when to speak, the most commonly used solution in those contexts is to use \textit{direct address}, that is, an explicit reference to the name of the chatbot in an utterance to command it to speak next: \textit{``Siri, what is next entry in my calendar?''} However, this makes interacting with the chatbots mechanical, less social, and in many situations, plainly awkward. Similarly, there are also situations in which multiple people may want to interact with a chatbot at the same time, as in a chat group, to coordinate among themselves and achieve a common goal. Also, some experimental conversational systems propose scenarios where one person interacts with several chatbots in the same conversation in order to compare or coordinate the services provided by them \cite{finch2018}. 

By avoiding direct address, chatbots need to know their proper turn to interact, which leads to what is referred to as the \textit{multi-party turn taking problem} \cite{Sacks1974}. Fundamentally, the goal is \textit{to predict which agent in the conversation is the most likely to speak next} and, conversely, when an agent must wait before interacting. An \textbf{\textit{agent}} can be either a person or a chatbot. That interaction can be a reply to the last interaction, a reply to an interaction in the past in the dialogue, or even a new idea or an interruption.  From the state of the art on conversational systems based only on text, we can find efforts, such as \textit{finch}, a multi-party system enabling interactions between people and four chatbots which are experts in financial investments, where turn-taking is controlled by a rule-based service which is called for every utterance  exchanged in the group chat, by considering both the content and the history of interaction between participants \cite{finch2018}. Even though the rule-based system solves the turn-taking problem for \textit{finch} in that specific domain, the approach presents limitations for scaling up the set of rules and the application to new domains since it is heavily dependent on expert's knowledge.

Therefore, the main contribution of this paper is to present and evaluate turn-taking as a machine learning (ML) problem, which can provide a more scalable solution. This involves defining a way to model turn-taking using a ML approach, and designing data sets to train and evaluate such models. For the former, given a finite set of possible agents that can speak, the algorithm tries to predict only the most likely agent to speak next, assuming only one agent should speak at a time, and the information to predict that can include only participant or both participant and content data. And for the latter, we present three corpora: one based on dialogues from seasons of an American TV situated comedy (sitcom), one with data gathered from the \textit{finch} system, and \textit{multibotwoz} created from the MultiWOZ dataset.

For validating the proposed approach, we have evaluated architectures with vanilla algorithms as \textit{Maximum Likelihood Estimation (MLE)} \cite{Harris98}, \textit{Support Vector Machines (SVMs)} \cite{Dogan2016} \cite{Guermeur2012}, \textit{Convolutional Neural Networks (CNN)} \cite{Collobert2018}, and \textit{Long-Short Term Memory (LSTM)} \cite{LiuJM15} to determine the most suitable technique for predicting turn-taking in conversations. 
We have found that the CNN models achieved higher accuracy than the other ML techniques on all three datasets, and that the content improved the overall performance of all approaches for the topic-oriented \textit{finch} and \textit{multibotwoz} dataset but not for the chit-chat dialogues of the sitcom dataset, where the ML models have not been able to beat the baseline. Finally, we found that the size of the corpus had a very positive impact on the accuracy for the content-based deep learning approaches, since we observe that those models perform best in the \textit{multibotwoz} dataset, i.e. the largest one.

\section{Related Work} 

To solve the task of managing the dialog of chatbots in a multi-party conversation, rule-based systems based on finite state automata have been proposed \cite{Emas18} \cite{Deepdial18}. However, in those works ML-based turn-taking models have not been explored, which can be an issue since rule-based systems tend to not scale well, specially considering that natural language evolves very quickly. 

On the other hand, several ML-based end-to-end data-driven dialogue systems have been built and evaluated \cite{Serban18}, including some which consider multi-party dialogues, albeit disentangling them into dyadic dialogues \cite{Elsner2008}. Further studies have also been conducted in order to build participant social role models \cite{Shaikh2010MPCAM}.

A related problem happens when chatbots are built as a collection of multiple chatbots with different skills, and after an utterance of the user the chatbot has to select which of the component chatbots must answer. In fact, this was the approach used by the majority of the contestants of the \textit{Alexa Prize}\footnote{Amazon's \textit{Alexa Prize}. https://developer.amazon.com/alexaprize} proposed by \textit{Amazon} to advance voice control skills and technologies. The 2018 chatbot winner implemented an open-domain dyadic social conversation system that achieved an average rating of 3.56 out of 5 in the last seven days of the semi-finals \cite{Alexa2018}. However, although these ML-based related works are extremely important for building better end-to-end dialogue systems, as far as we know, none of them have been employed to learn multi-party turn-taking models from conversation logs with the goal of predicting who speaks next.

 The most closest work is \cite{Ouchi2016}, in which a model that encoded the context to predict the addressee and a response in multi-party conversation was proposed. We, on the other hand, do not try to predict the response. Furthermore, there are datasets as \textit{Reddit} or \textit{Ubuntu IRC} \cite{Uthus2013} \cite{Lowe15} which are forum-oriented and they do not have the structure of a conversation with clear speaker turns. We have studied the transformation of the Ubuntu IRC dataset into a multi-party dialogue dataset. However, it is far from trivial. The original training corpus is organized in days with no clear threads of conversation between the participants. Therefore, there are multiple conversations happening concurrently each day. After applying some filters, we realized that this effort requires sophisticated algorithms and is out of the scope of our paper.
 
Related to \textit{finch} dataset is the Multi-Domain Wizard-of-Oz dataset (MultiWOZ)\footnote{MultiWoZ dataset: https://www.repository.cam.ac.uk/handle/1810/280608} \cite{Budzianowski18}, which is a fully-labeled collection of human-human written conversations spanning over multiple domains and topics. However, this dataset was not created considering that more than one bot would be in the conversation with the user. Rather, it was created considering a dyadic conversation between the user and a bot that can talk about multiple domains or topics. The topics are actually service providers. We have though adapted this dataset as explained in next section.

\section{Datasets} 
\label{sec3}

For learning multi-party turn-taking models we considered three datasets: a corpus based on dialogues from scripts of 10 seasons of a very popular American sitcom; the \textit{finch} corpus with logs of real-world interactions between people and four chatbots \cite{finch2018}, and \textit{multibotwoz}\footnote{Multibotwoz dataset: to be provided by the authors upon request.} which we created as an adaptation from the MultiWoZ dataset \cite{Budzianowski18}, a multi-domain service-based dialogue corpora. \textit{finch} was designed so that the chatbots cooperate to advise the user (a person) to choose among three investment options: savings account, certificate of deposit, and treasure bonds. 

For each of those investment options there is an expert chatbot (\textit{saGuru}, \textit{cdGuru} and \textit{tdGuru}, respectively) plus a fourth chatbot (\textit{inGuru}) which is a mediator moderating the conversation and assuring that the user's utterances are replied.  The expert chatbots are not only able to give answers related to investments but can also estimate the return on investment for an initial amount and period of time specified by the user. 

In general, the American sitcom dataset is more chit-chat oriented, while the \textit{finch} corpus contains human-machine interactions and is more topic-oriented in a way that each chatbot interacts only depending on the topic. For the American sitcom corpus, we considered each scene of each episode as a single dialogue, and considered only the scenes containing at least three (3) and at most six (6) agents (considering only the six main characters of the sitcom). The actual corpus ended up with $20,086$ utterances in $1,050$ dialogues with $4$ agents on average in each dialogue and $19$ utterances exchanged on average in each dialogue. 

Regarding the \textit{multibotwoz} dataset, we have filtered the MultiWoZ dataset with the multi-domain-based dialogues only, i.e., the dialogues that the user requested more than one service. And we have filtered (within this subset) the ones that contained only the following services: attractions, hotel, restaurant, taxi and train. We did that because the amount of dialogues were not sufficient as a sample to learn turn taking for these services. We then updated the bots name depending on the service (one bot for each service). To perform this task, we have created a pool of classifiers that based on the domain-based dictionaries and the classification of the two last utterances sent in the dialog, could determine the domain of a given utterance. 

\begin{table*}[hbt]
\centering
\begin{tabular}{@{}rrrrcrrr@{}}
\toprule 
 \textbf{Metric} & \textbf{sitcom}  & \textbf{finch} & \textbf{multibotwoz}    
\\
\midrule
\textit{Corpus size (number of utterances)} & 20,086	& 1,148 & 99,553 \\
\textit{Total number of Dialogues} & 1,050	& 41 & 6,138 \\ 
\textit{Avg. number of agents per Dialogue} &  4 & 5 & 4 \\ 
\textit{Avg. number of utterances per Dialogue} & 19 & 24 & 16 \\
\textit{Avg. length of utterances (words)} & 11 & 12 & 13\\
\bottomrule
\end{tabular}
\caption{Datasets comparison - Summary.}
\label{tab:corpus}
\end{table*}

\begin{table*}[hbt]
\centering
\small
\begin{tabular}{@{}rrrrrrp{2.5cm}rrr@{}}
\toprule
&
 \multicolumn{2}{p{2cm}}{\textbf{sitcom}} & \multicolumn{2}{p{2cm}}{\textbf{\textit{finch}}}  & \multicolumn{2}{p{2cm}}{\textbf{\textit{multibotwoz}}}  
\\
\midrule
 $Agent$ & $\% interactions$ & $Agent$ & $\% interactions$ & $Agent$ & $\% interactions$ 
\\
\midrule
 \textit{no user} & - & \textit{finch-user} & 38\% & \textit{user} & 50\% \\
\textit{A} & 18\% & \textit{inGuru} & 33\% & \textit{train\_bot} & 12\%  \\
\textit{B} & 18\% & \textit{cdGuru} &  12\% & \textit{hotel\_bot} &  12\% \\ 
\textit{C} &  16\% & \textit{tdGuru} &  10\%  & \textit{restaurant\_bot} & 11\% \\
\textit{D} & 16\% & \textit{saGuru} &  7\%& \textit{attraction\_bot} & 10\%\\ 
\textit{E} & 16\% & - & - & \textit{system\_bot} & 4\% \\ 
\textit{F} & 16\% & - & - & \textit{taxi\_bot} &  2\% 
\\
\bottomrule
\end{tabular}
\center
\caption{Datasets comparison - Distribution of Utterances per Agent.}
\label{tab:distagent}
\end{table*}

\begin{table}[hbt]
\centering
\small
\begin{tabular}{@{}rrrrcrrrr@{}}
\toprule 
  & \textbf{\textit{A}} & \textbf{\textit{B}} & \textbf{\textit{C}} & \textbf{\textit{D}} & \textbf{\textit{E}} & \textbf{\textit{F}} 
\\
\midrule
\textbf{\textit{A}} & - &      19.52 &      18.73 &      16.47 &      23.37 &      19.95 \\ 
\textbf{\textit{B}} &      20.93 & - &      20.62 &      25.91 &      13.62 &      17.40 \\ 
\textbf{\textit{C}} &      15.84 &      21.74 & - &      20.12 &      27.64 &      12.80 \\ 
\textbf{\textit{D}} &      18.63 &      \textbf{28.68} &      21.63 & - &      \textbf{12.77} &      16.96 \\ 
\textbf{\textit{E}} &      24.59 &      13.95 &      26.22 &      15.18 & - &      18.56 \\ 
\textbf{\textit{F}} &      24.13 &      20.70 &      15.45 &      17.35 &      19.90 & - \\
\bottomrule
\end{tabular}
\caption{Interaction frequency between agents (\%) in the American sitcom dataset.}
\label{tab:freqsitcom}
\end{table}
\begin{table}[h]
\centering
\small
\begin{tabular}{@{}rrrrcr@{}}
\toprule 
 & \textbf{\textit{user}} & \textbf{\textit{saGuru}} & \textbf{\textit{tdGuru}} & \textbf{\textit{cdGuru}} & \textbf{\textit{inGuru}} \\  
\midrule
\textbf{\textit{user} }  & - &     8.42 &    16.84 &    23.68 &    51.05 \\ 
\textbf{\textit{saGuru} } &    23.68 & - &    23.68 &    39.47 &    13.15 \\ 
\textbf{\textit{tdGuru}} &    47.54 &     6.55 & - &    14.75 &    31.14 \\ 
\textbf{\textit{cdGuru}} &    60.27 &     6.84 &    19.17 & - &    13.69 \\ 
\textbf{\textit{inGuru}}  &    \textbf{79.85} &    12.23 &     5.03 &     \textbf{2.87} & - \\ 
\bottomrule
\end{tabular}
\caption{Interaction frequency between agents (\%) in the \textit{finch} dataset.}
\label{tab:freqFinch}
\end{table}

\begin{table}[hbt]
\centering
\small
\begin{tabular}{@{}rrrrcrrr@{}}
\toprule 
 \textbf{Agent} & \textbf{Interaction frequency after \textit{user}(\%)}   \\
\midrule
\textit{train\_bot} & \textbf{25.2587}\%  \\ 
\textit{hotel\_bot} &   23.2920\%  \\ 
\textit{attraction\_bot} & 20.3226\%  \\ 
\textit{restaurant\_bot} & 21.5631\%  \\ 
\textit{system\_bot} & 4.7304\%  \\ 
\textit{taxi\_bot} &  \textbf{4.8332}\% \\ 
\bottomrule
\end{tabular}
\caption{Interaction frequency in the multibotwoz dataset.}
\label{tab:freqcamb}
\end{table}

For the sentences that were related to generic dialog acts\footnote{Also known as speech acts and represent the function of the speech.}, as greetings and thanks, or clarification questions, we have defined that these utterances were sent by the \textit{system\_bot}. It then played the role of a mediator in the conversation, just like \textit{inGuru} bot does in \textit{finch} dataset. The resulted corpus ended up with $99,556$ utterances in $6,138$ dialogues with $4$ agents on average in each dialogue, varying from $3$ to $8$ and $16$ utterances exchanged on average in each dialogue. 

A summary of the the three corpora is listed in Tables \ref{tab:corpus} and \ref{tab:distagent}. The \textit{finch} corpus is much smaller than the sitcom and \textit{multibotwoz} corpora, comprising  $1,148$ utterance exchanges in a total of $41$ dialogues, with a $5$ average number of agents and $24$ utterances exchanged per dialogue on average. Despite that, it was created in a real-world system in which people interacted with multiple bots in the same conversation. Further, we present in Tables \ref{tab:freqsitcom}, \ref{tab:freqFinch} and \ref{tab:freqcamb} the interaction frequencies between agents, to better characterize the difference between the three datasets. Element ${a_i}_j$ represents the frequency of agent $j$ interacting after agent $i$. Note that we did not compute the frequency for the same agent. In all datasets we have merged consecutive utterances by the same agent into one utterance.

In the sitcom dataset the interaction frequencies between agents are reasonably similar, as expected in typical chit-chat scenarios, with the minimum value of the interaction frequency between agents being $12.77\%$ (agent E after D) and the maximum value $28.68\%$ (agent B after D),  with an average of $19.65\%$ and standard deviation of $4.33$.  However, the same does not happen in the \textit{finch} and \textit{multibotwoz} datasets. The minimum value in the former is $2.87\%$ (\textit{cdGuru} after \textit{inGuru}) and the maximum value is $79.85\%$ (\textit{finch-user} after \textit{inGuru}) with average of $24.99\%$ and standard deviation of $20.73$. While in the later, minimum value is $4.83$ for \textit{taxi\_bot}, and the maximum value is $25.26$ for \textit{train\_bot}. This clearly shows that there is more variability on the conditional interaction for the dialogues in these ones than for the sitcom. For instance, this makes easier to accurately predict in \textit{finch} when the \textit{user} interacts after \textit{inGuru}, and similarly that it is very uncommon for \textit{cdGuru} to speak after \textit{inGuru}. Finally, no bot interacts after another bot in the \textit{multibotwoz} dataset, only after the \textit{user} with frequencies illustrated in table \ref{tab:freqcamb}.

\section{Machine Learning-based Models}
In this section we describe the ML-based methods which we have considered in this research. Briefly, we have evaluated different types of ML approaches, such as MLE, SVMs, and Neural Networks, and for some of the approaches we have varied the implementation by considering only \textit{agent} information, i.e. only who spoke the previous utterances; or by taking into account also the \textit{content} in of the utterances, i.e. who spoke and what was said. In addition, the methods differ in the way the agent information is encoded. For most methods, a binary-based encoding is considered, while for the Neural Networks the agents are encoded as raw text. We refer to these methods as \textit{Traditional ML} and \textit{Deep Learning} Methods, respectively.

\subsection{Baselines}

We consider a baseline which we call \textbf{Repeat Last} to compare our proposed methods: this approach is based on a social rule often observed in multi-party human dialogues \cite{Sacks1974}: when more than two agents are exchanging utterances, there is a tendency that the agent speaking before the current to be the SNS. Thus, whenever an agent speaks, we might predict the next one as being the one that had spoken before.  More formally, the \textit{Repeat Last} baseline prediction works as following: let $A=\{a_i  |  1 \leq i \leq n\}$ be the set of agents in the dialogue, $n$ be the number of agents, and let $S = \{s_t  |  1 \leq t \leq T\}$ be the set of agents who sent an utterance in the dialogue up to a time $T$, where $s_t \in A$. Whenever the speaker $s_t$ sends an utterance, the next agent selected to talk, denoted $s_{t+1}$, is the one who spoke at time $t-1$, i. e., $s_{t+1} = s_{t-1}$.

\begin{figure}[h!]
\centering
\begin{tabular}{@{}rcrcrcrcrcrcrcr@{}}
\toprule 
\small
 \textbf{Speaker} &  \small$s_{1}$ &  \small$s_{2}$ &  \small$s_{3}$ & \small$s_{4}$ & \small$s_{5}$ &  ... &   \small$s_{t}$ &  \small$s_{t+1}$  \\
\midrule
\small \textbf{Next Speaker}  & - &  - &  \small$s_{1}$  & \small$s_{2}$ & \small$s_{3}$ &  ... &  \small$s_{t-2}$ &  \small$s_{t-1}$  \\
\bottomrule
\end{tabular}
\caption{\textit{Repeat Last} Baseline - Example}
\label{fig:repeatLast}
\end{figure}

\subsection{Traditional ML Methods}
For the more traditional ML methods, such as MLE, SVM, and the like, we make use of the \textit{one-hot} encoding to convert the information of the agents to a feature vector, formalized as follows. Let $x$ be a vector and $x \in C^n $, a $n$-dimensional instance space with $n$ agents in the conversation and $a_i$ the $i$-th agent, where $1 \leq i \leq n$, and let $s_t$ be the agent who spoke at time $t$. The binary feature vector $x_t$ for predicting the next speaker at time $t+1$, in the simplest case with a lookback window with size equals to 1, can be defined as:
\begin{equation}
\label{eq:1}
 x(t) = 
 \begin{bmatrix} {x_{t}}_1 & {x_{t}}_2 & ... & {x_{t}}_n \end{bmatrix}^T  
 \end{equation}
 \begin{equation}
 {x_t}_i = 
 \begin{cases}
    1  & \quad \text{if } a_i \text{ is the sender, i.e. } a_i = s_t\\
    0  & \quad \text{if } a_i \text{ is not the sender, i.e.} a_i \ne s_t
  \end{cases}
\end{equation}
We perform a linear transformation $T : C^n \rightarrow C^{W * n}$ on $x(t)$, by taking into account $x(t)$ until $x(t-(W+1))$, where $W$ is the size of lookback window  and $t > W$, as:
\begin{equation}
\small
 x'(t) = 
 \begin{bmatrix} {x_{t}}_1 & \ldots & {x_{t}}_n  & {x_{t-(W+1)}}_1 &  \ldots & {x_{t-(W+1)}}_n \end{bmatrix}^T
 \label{eq:binary}
\end{equation}
\subsubsection{Agents-based models:}
The approaches considered herein encode only the information of the agents, by making use of the aforementioned one-hot encoding method. The methods are the following:\\

\noindent \textbf{A-MLE}: \textit{Maximum Likelihood Estimation} \cite{Harris98} taking into account only the order in which the agents interact in the conversation. Therefore, transitions are learned by considering that the previous state is the last agent which sent an utterance and the next state is the following agent which sent an utterance. We also modeled A-MLE considering a lookback window of size 2, which means the previous state contains information of the two last agents which sent an utterance. In this case, a $\theta$ transition from state $\pi-1$ to $\pi$, is modeled as:
\begin{equation}
\label{eq:bmle}
\theta : state(\pi-1) =  x'(t) \rightarrow state(\pi) = x(t)
\end{equation}
We then compute the MLE with smoothing to estimate the parameter for each $\theta (\pi) \in \Theta$ transition type. Therefore, for each corpus, we estimate $L$ for observed transitions as:
\begin{equation}
\label{eq-likelihood}
 L (\theta | x'(t), x(t))= 
 \frac{\displaystyle count(\theta,x'(t), x(t))+1}{\displaystyle count(\theta,x'(t), x(t))+|\Pi|}
\end{equation}
Where $\Pi$ is the set of states and $|\Pi|$ is the number of states in the set.\\

\noindent \textbf{A-SVM}: a multi-class linear \textit{SVM} model \cite{Dogan2016} is trained with $n$ classes: one for each agent. The model is trained by receiving as input binary vectors of length $n$, like in A-MLE.  However, in this case, a class is predicted considering:
\begin{equation}
x'(t) \rightarrow \ell(x(t)) | \ell \in L
\end{equation}
where $L$ is the label set which contains the names of the agents.\\

\noindent \textbf{BA-SVM}: A binary \textit{SVM} model is learned for each agent, which can classify whether the respective agent is likely to reply or not for a given utterance. The dialogues are parsed and only the agent encoded vectors are considered. Then, for each utterance, if the agent is the speaker of the utterance, then the agent's name is assigned as the label, otherwise the \textit{"Other"} label is assigned.  The training data are used as input in a SVM modeler and the generated models are saved.  For each example in the testing data, all models are called and the output is ranked. The top 1 of the ranking list is chosen as the next most likely speaker. 

\begin{figure}[h]
\centerline{\includegraphics[width=8cm]{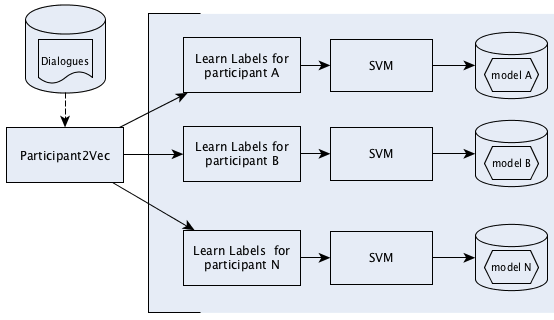}}
\caption{\small{BA-SVM Learning Architecture.}}
\label{fig:bp-svm-arch}
\end{figure}

\subsubsection{Agents-and-Content-based Models:}
Below we describe variations of the previously described methods but with the addition of content information, that is, what was actually spoken by each agent.\\

\begin{figure}[hbt]
\centerline{\includegraphics[width=8.5cm]{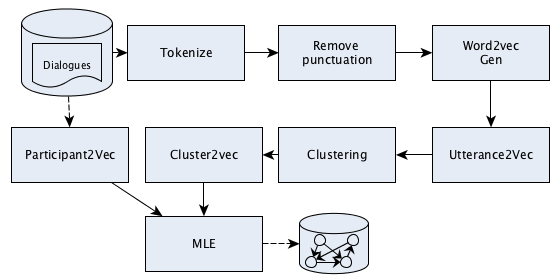}}
\caption{\small{AC-MLE Learning Architecture.}}
\label{fig:mle-arch}
\end{figure}

\begin{figure}[hbt]
\centerline{\includegraphics[width=7cm]{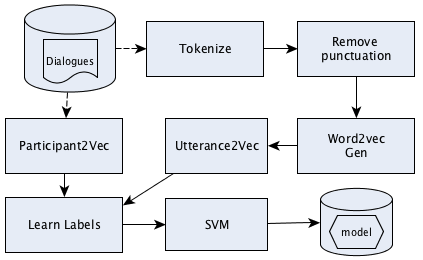}}
\caption{\small{AC-SVM Learning Architecture.}}
\label{fig:svm-arch}
\end{figure}

   \noindent \textbf{AC-MLE}: The agent-and-content MLE-based architecture considers also the utterance which is being exchanged in the dialogue in addition to the agent vector as defined in Equation~\ref{eq:bmle}. The utterances are first tokenized and punctuation symbols are removed, having a lists of tokens as result. By taking into account all utterances in the training set, a \textit{Word2Vec} embedding model is trained. We have used the \textit{gensim}\footnote{https://radimrehurek.com/gensim/models/word2vec.html} library for generating the word2vec models. After this process, the feature vector (\textit{Utterance2Vec}) representing the utterance consists of the mean of all of its corresponding word vectors. Then, the utterance vectors are given as input to a \textit{K-means} clustering algorithm \cite{Macqueen67}. 

A binary vector which represents the detected cluster for each utterance is generated as well as a binary vector which represents the agent which sent the utterance, both considering the one-hot encoding method. Then, the cluster and agent vector pairs are used to train a MLE-based model. More formally, let $x'(t) $ be a vector as in Equation \ref{eq:binary}, but in this case concatenated with the binary vector described in the previous paragraph, then a class is predicted by computing the transitions as in Equation ~\ref{eq:bmle} and the likelihood as in Equation ~\ref{eq-likelihood}.\\

 \noindent \textbf{AC-SVM}: This agent-and-content SVM-based architecture makes use of word embeddings for better capturing the semantic meaning of the utterances. 

Then a multi-class linear SVM model is trained also with $n$ classes as in the A-SVM approach, however it receives as input both utterance and agent vectors concatenated into single vectors. More formally, let $x$ be a vector as in Equation \ref{eq:1} and $u$ be the utterance vector. We perform a linear transformation $T : C^n \rightarrow C^{W*n+|u|}$ on $x(t)$ by taking the $x(t-W)$, where $W$ is the lookback window. So, for $W=1$, we have: 
\begin{equation}
\label{eqsvm}
\small
 x'(t) = 
 \begin{bmatrix} {x_{t-1}}_1 & ... & {x_{t-1}}_n  & {u_{t-1}}_1 &  ... & {u_{t-1}}_n \end{bmatrix}^T
\end{equation}
Then a class is predicted considering:

\begin{equation}
x'(t) \rightarrow \ell(x(t)) | \ell \in L
\end{equation}

Where $L$ is the label set which contains the names of the agents.

\subsection{Deep Learning Methods}
We consider two different deep learning methods. The first is based on a CNN \cite{Collobert2018} which is a model generally applied on classification tasks, and the second is based on LSTM \cite{LiuJM15} which is a model also generally used for classification tasks but with an extra capability of learning temporal information which is particularly attractive in the context of dialogues. \\

 \noindent \textbf{AC-CNN}: The agent-and-content \textit{convolutional neural network (AC-CNN)} presented herein consists of a standard model used for text classification adapted for the task involved in this paper. Such adaptation consists of formatting the previous utterances and the name of the agent as a raw text, and defining the label as in the previous methods. 
 
 \begin{figure}[h]
\centerline{\includegraphics[width=\textwidth]{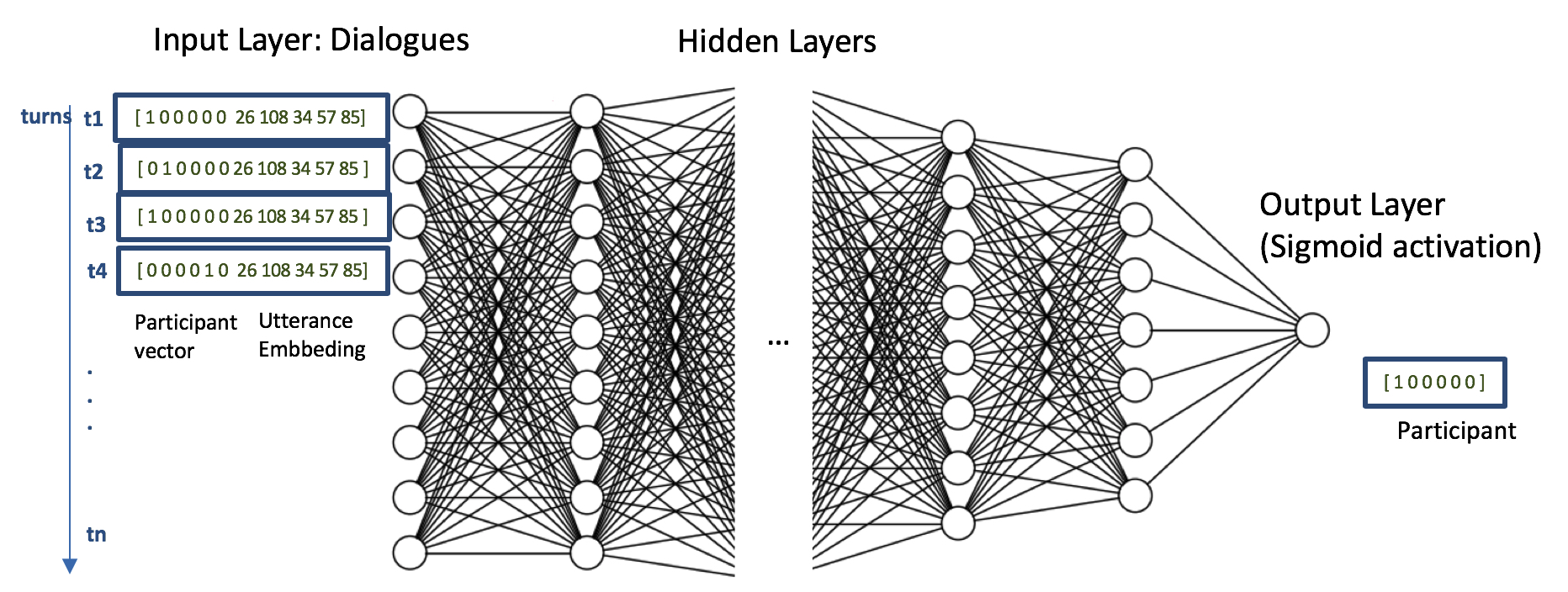}}
\caption{\small{AC-CNN Learning Architecture.}}
\label{fig:pc-cnn}
\end{figure}
   
   More formally, let $s_{t-1}$ be the agent who spoke utterance $u_{t-1}$ at time $t-1$, and $s_t$ the agent who spoke the last utterance $u_{t}$, to predict who will speak at time $t+1$, we build the following raw text: $s_{t-1} \oplus u_{t-1} \oplus s_t \oplus u_t$, where $\oplus$ represents the concatenation of textual strings. That text is then used as input to the neural network.

The architecture considered for the CNN is the following: embedding layer with 64 dimensions; dropout set to 0.2; convolutional layer with 64 filters with kernel size of 3 and stride equals to 1; 1D Global Max-pooling layer with pool size set to 5; another dropout set to 0.2; and 300-dimensional dense hidden layer. \\

 \noindent \textbf{AC-LSTM}: For the agent-and-content long-short term memory neural network, we make use of the same raw-text-based encoding we described for the CNNs, and the main difference lies in the architecture of the two methods, since LSTMs contain a layer aiming at learning temporal information. Given that a dialogue consists of a sequence of pairs comprised of a speaker and an utterance, the goal of evaluating this model is to investigate whether such temporal sequence can be captured and learned by a ML model.

\begin{figure}[h]
\centerline{\includegraphics[width=11cm]{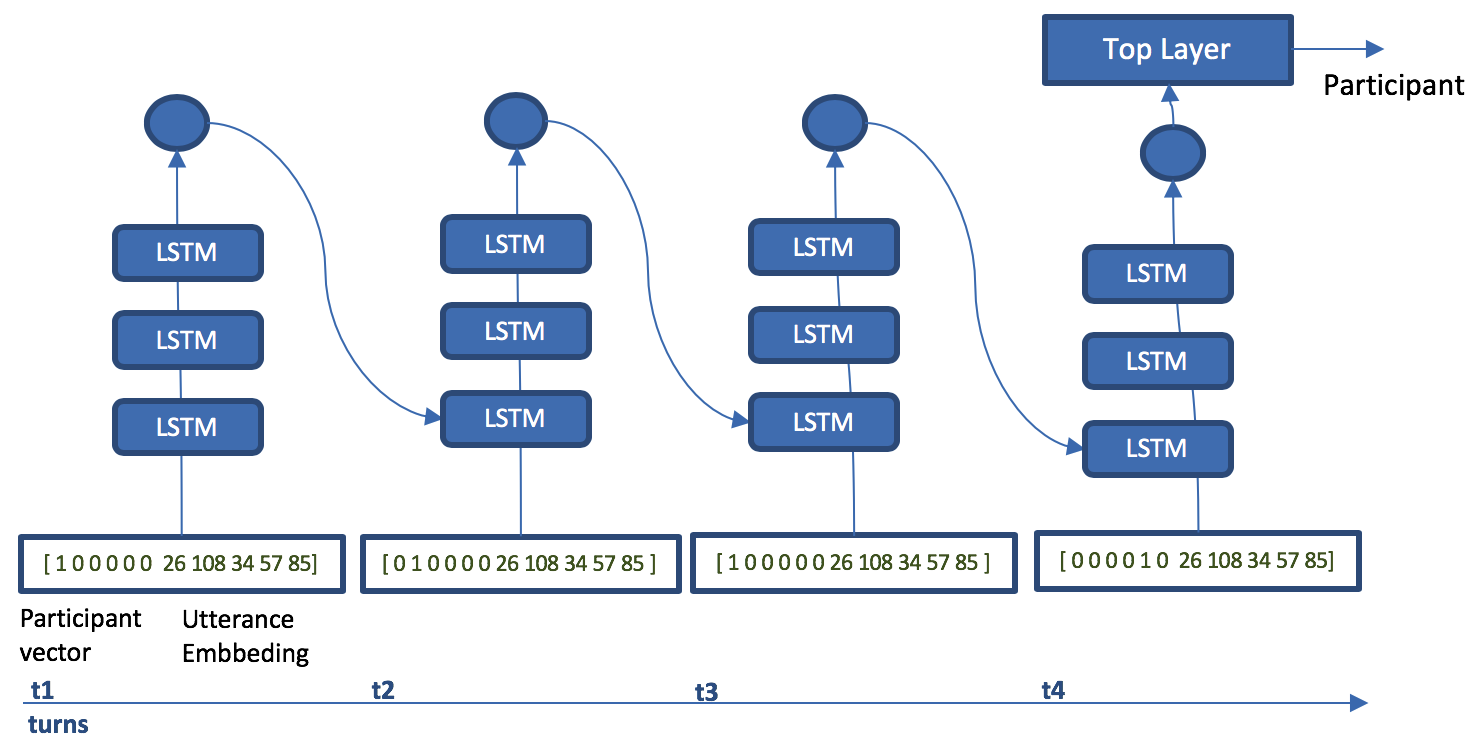}}
\caption{\small{AC-LSTM Learning Architecture.}}
\label{fig:pc-lstm}
\end{figure}

The architecture we considered for this neural network is the following: embedding layer with 64 dimensions; dropout set to 0.25; convolutional layer with 64 filters with kernel size of 3 and stride equals to 1; and 1D Global Max-pooling layer with pool size set to 5. For this model we have set meta-parameters that are similar to that of the CNN, being the only exception the number of epochs which was set to 2 for the LSTM. By removing the text from the input, we have also implemented two variations of the neural networks only with agent information to compare with the agent-only models: (i) {A-CNN}: the same architecture as AC-CNN; (ii) {A-LSTM}: equivalent to AC-LSTM.
\section{Evaluation Results} 
In this section we present the training approach and the results with the agent-based baselines models and the learning models considering both content and agent information, for both the sitcom dataset and the \textit{finch} dataset.

\subsection{Training Approach}

Our model does not constraint with regard to waiting for a specific moment to start predicting, it follows a more classical batch-learning process. For both corpora, we considered a $70/30$ train-test split, where $70\%$ of subsequent dialogues are used for training and the remaining $30\%$ for testing. In order to set meta-parameters for the models, cross-validation has been applied on the training set. Regarding the number of clusters for the MLE-based architecture, after some trials and observing how the clusters were created in a PCA \cite{WOLD198737} 2D plot, we parametrized the model with six (6) clusters for the sitcom data, five (5) for \textit{finch} data and seven (7) for \textit{multibotwoz}. Although the number of classes are identical to the number of agents, we did not find a correlation between the clusters and the agents or the conditional interaction between the agents. The vocabulary is built with training and testing data, therefore, all words had WE and there were no words which where OOV. For both the embedding and the hidden layers in the AC-CNN models, \textit{Rectified-Linear-Units activation functions (Relu)} are applied. For the training, we make use of the \textit{Adam optimizer}, with 3 epochs for training and learning rate set to 0.001. Batch size is set to 50 for the sitcom dataset, and 5 for {\it finch} and {\it multiwoz} data. And for the AC-LSTM architecture, one LSTM layer was considered with output size set to 90 for the TV sitcom dataset and to 50 for the {\it finch} and {\it multiwoz} datasets. To evaluate the models and compare the results, we have computed the accuracy metric.  

\subsection{Evaluation of the Proposed Methods}

\begin{table}[b!]
\centering
\small
\begin{tabular}{@{}rrrrrrrp{0.5cm}@{}}
\toprule
&
 \multicolumn{2}{p{0.5cm}}{\textbf{sitcom}} & \multicolumn{2}{p{0.5cm}}{\textbf{\textit{finch}}}  & \multicolumn{2}{p{0.5cm}}{\textbf{\textit{multibotwoz}}}  
\\
\textit{Baseline} &  \multicolumn{2}{p{0.5cm}}{61.34} & \multicolumn{2}{p{0.5cm}}{57.64}  & \multicolumn{2}{p{0.5cm}}{86.49}  
\\
\midrule
& $W_1$ & $W_2$ & $W_1$ & $W_2$ & $W_1$ & $W_2$ 
\\
\midrule
\textit{A-MLE}   & 26.75  & 57.61  & 57.89  & \textbf{67.76} & 62.26 & 83.06
\\
\textit{A-SVM}   &  26.78 & 57.69  & 66.88 & 66.55 & 62.26 & 83.06
\\
\textit{BA-SVM}  & 17.48  & 57.82   & 56.39 & 61.96 & 62.26 & 83.06
\\
\textit{A-CNN}   & 24.81  & \textbf{61.34} & 60.79 & 63.54 & 59.70 & \textbf{86.38}
\\
\textit{A-LSTM}   & 17.05 & 13.40 & 38.54 & 36.11  & 46.66 & 50.00
\\
\bottomrule
\end{tabular}
\center
\caption{Accuracy for models considering only the agents as input. (\textit{W: lookback window}).}
\label{tab:acc}
\end{table}

\begin{table}[b!]
\centering
\small
\begin{tabular}{@{}rrrrrrrp{0.5cm}@{}}
\toprule
&
 \multicolumn{2}{p{0.5cm}}{\textbf{sitcom}} & \multicolumn{2}{p{0.5cm}}{\textbf{\textit{finch}}}  & \multicolumn{2}{p{0.5cm}}{\textbf{\textit{multibotwoz}}} 
\\
\textit{Baseline} &  \multicolumn{2}{p{0.5cm}}{61.34} & \multicolumn{2}{p{0.5cm}}{57.64}  & \multicolumn{2}{p{0.5cm}}{86.49}  
\\
\midrule
& $W_1$ & $W_2$ & $W_1$ & $W_2$ & $W_1$ & $W_2$ 
\\
\midrule
\textit{AC-MLE}   & 26.65 & 57.68  & 64.47 &  67.76 & 74.41 & 86.34
\\
\textit{AC-SVM}   & 26.75  & 57.67  & 58.03 & 65.24 & 86.84 & 92.48
\\
\textit{AC-CNN}     & 30.77 & \textbf{61.63} & 66.77 & \textbf{69.44}& 86.03 & \textbf{94.19}
\\
\textit{AC-LSTM}     & 23.07 & 60.60 & 63.45 & 63.88 & 83.41 & 93.31
\\
\bottomrule
\end{tabular}
\center
\caption{Accuracy considering both agents and content information. (\textit{W: lookback window}).}
\label{tab:acc2}
\end{table}

\begin{table}[b!]
\centering
\begin{tabular}{@{}rrrrc@{}}
\toprule
& \textbf{sitcom} & \textbf{\textit{finch}} & \textbf{\textit{multibotwoz}} 
\\
\midrule
\textit{A-MLE}   & -3.73 & \textbf{10.14} & -3.43
\\
\textit{A-SVM}   & -3.65 & 8.93 & -3.43
\\
\textit{BA-SVM}  & -3.52 & 4.34 & -3.43
\\
\textit{A-CNN}  &  \textbf{0.00}  & 5.90 & -26.49
\\
\textit{A-LSTM}  &  -47.94 & -21.53 & -0.11
\\
\midrule
\textit{AC-MLE}   & -3.66 &	\textbf{10.14} & -0.10
\\
\textit{AC-SVM}   & -3.67 & 7.62 &  5.99
\\
\textit{AC-CNN}      &  \textbf{0.29} & \textbf{11.80} & \textbf{7.70}
\\
\textit{AC-LSTM}     &  -0.74 & 6.24 & 6.82
\\
\bottomrule
\end{tabular}
\center
\caption{Difference in percentage points between \textit{Repeat Last} baseline and results from Table \ref{tab:acc2} with W=2.}
\label{tab:acc-diff}
\end{table}

The \textit{Repeat Last} baseline achieves an accuracy of $61.34\%$ on the sitcom dataset, $57.64\%$ on \textit{finch}, and $86.49\%$ on \textit{multibotwoz}. From the results shown in Tables \ref{tab:acc} and \ref{tab:acc2} we can see in all datasets that higher accuracy values can be achieved when two turns are considered, i.e. with a lookback window equals to two, than when only the last turn is considered. Although we have also trained MLE-based models with lookback windows ranging from three to five, we do not present the results here because they did not improve over the accuracy of the models with lookback window equals to two. 

In general, almost all models performed equal or below the baseline for the sitcom dataset, while much greater improvements in accuracy were seen in the \textit{finch} dataset for several of the models. Regarding the \textit{multibotwoz} dataset, the models that used only the agent information could not beat the baseline, while CNN actually had the lowest performance. On the other and, the agent and content based CNN model had the best performance compared to all others models and datasets ($94.19\%$). We present in table \ref{tab:acc-diff} the difference of percentage points between \textit{Repeat Last} baseline and results with lookback window equals to two. Unfortunetly, a chatbot can not determine which approach to consider (baseline or learned model) based on the type of conversation (chit-chat or topic-oriented) and the size of the dialog with these results, because the results for the sitcom dataset compared to the baseline were not statistical significant ($p < 0.01) $. However, the results show: (i) the size of the corpus has a very positive impact on the accuracy for the content-based deep learning approaches and those models perform best in the larger datasets, since the results were statistical significant over the baseline for the \textit{multibotwoz} dataset ($p < 0.01$); and (ii) if the dialogue dataset is small and topic-oriented (but with few topics), which is the case of \textit{finch} dataset, it is sufficient to use an agent-only MLE or SVM models, although slightly higher accuracies can be achieved with the use of the content of the utterances with a CNN model ($p < 0.01$).

\section{Conclusions and Future Work} 

In this paper we investigated the application of machine learning (ML) techniques to  learn multi-party turn-taking models from three dialogue corpora.

We presented results which indicate that if the dialogue dataset is small and topic-oriented (but with few topics), it might be sufficient to use an agent-only MLE or SVM models, although slightly higher accuracies can be achieved with the use of the content of the utterances with a CNN model. However, if the dialogue dataset is bigger, our results indicate that an agent-and-content  CNN model performs best, albeit almost at the level of a very simple, baseline model which simply uses the speaker before the last as its prediction. 

Further studies could be done in order to find the best number of clusters for the clustering algorithm in the AC-MLE approach. As future work, online and reinforcement learning could also be tested, so a chatbot would be able to learn turn-taking during interaction, enabling a self-adaptive behavior on the turn-taking model. Finally, research on syntactic variations of the speaker to be used for prediction could be done.

\bibliographystyle{unsrt}
\bibliography{arxiv-multiparty19-bibliography}

\label{lastpage}
\end{document}